\documentclass{bmvc2k}
\usepackage{amsmath,amssymb,wrapfig}
\usepackage{graphicx}
\makeatletter
\newcommand{\printfnsymbol}[1]{%
  \textsuperscript{\@fnsymbol{#1}}%
}

\title{PrOSe: Product of Orthogonal Spheres Parameterization for Disentangled Representation Learning}

\addauthor{Ankita Shukla}{ankitas@iiitd.ac.in}{1}
\addauthor{Sarthak Bhagat*}{sarthak16189@iiitd.ac.in}{1}
\addauthor{Shagun Uppal *}{shagun16088@iiitd.ac.in }{1}
\addauthor{Saket Anand}{anands@iiitd.ac.in}{1}
\addauthor{Pavan Turaga}{pturaga@asu.edu}{2}
\addinstitution{
IIIT - Delhi\\
Delhi, India
}
\addinstitution{
 Arizona State University.\\
 Arizona, USA
}

\runninghead{Shukla et al:}{Product of Orthogonal Spheres Parametrization}

\def\etal{\emph{et al}\bmvaOneDot}

\newcommand{\bbR}{\mathbb{R}}

\newcommand{\calS}{\mathcal{S}}
\newcommand{\calX}{\mathcal{X}}
\newcommand{\calL}{\mathcal{L}}

\newcommand{\bx}{\mbox{$\mathbf{x}$}}
\newcommand{\bz}{\mbox{$\mathbf{z}$}}
\newcommand{\bA}{\mbox{$\mathbf{A}$}}
\newcommand{\bJ}{\mbox{$\mathbf{J}$}}
\newcommand{\bI}{\mbox{$\mathbf{I}$}}
\newcommand{\bZ}{\mbox{$\mathbf{Z}$}}
\newcommand{\bU}{\mbox{$\mathbf{U}$}}

\begin{document}

\maketitle

\begin{abstract}
Learning representations that can disentangle explanatory attributes underlying the data improves interpretabilty as well as provides control on data generation. Various learning frameworks such as VAEs, GANs and auto-encoders have been used in the literature to learn such representations. Most often, the latent space is constrained to a partitioned representation or structured by a prior to impose disentangling. In this work, we advance the use of a latent representation based on a product space of Orthogonal Spheres \textit{\textbf{PrOSe}}. The PrOSe model is motivated by the reasoning that latent-variables related to the physics of image-formation can under certain relaxed assumptions lead to spherical-spaces. Orthogonality between the spheres is motivated via physical independence models. Imposing the orthogonal-sphere constraint is much simpler than other complicated physical models, is fairly general and flexible, and extensible beyond the factors used to motivate its development. Under further relaxed assumptions of equal-sized latent blocks per factor, the constraint can be written down in closed form as an ortho-normality term in the loss function. We show that our approach improves the quality of disentanglement significantly. We find consistent improvement in disentanglement compared to several state-of-the-art approaches, across several benchmarks and metrics. 
\end{abstract}
{\let\thefootnote\relax\footnote{{* equal contribution}}}
\section{Introduction}\label{sec: intro}
Representation learning \cite{higgins2018towards,Bengio2013RepresentationLA} approaches that can mimic human understanding of different attributes has increased in attention in recent years, often referred to as disentangled representations. These representations allow to separate various explanatory factors underlying the data. Disentangled representations are easily interpretable when they align with the true data generative factors. However, inverting the generative process is extremely challenging owing to the complex interaction between different attributes. Therefore, over the last decade several approaches have been developed that utilize either generative model variants \cite{Mathieu2016DisentanglingFO, Matthey2017betaVAELB, Jha2018DisentanglingFO, Szab2018ChallengesID} or autoencoder variants \cite{Mixing_2018,Li2019} to learn such representations.

Approaches that focus on imposing variable level independence, have been successful by using a modification of VAE framework. Recent works like $\beta$-VAE \cite{Matthey2017betaVAELB}, Factor-VAE \cite{Kim2018DisentanglingBF} fall under this category. On the other hand, several other approaches partition the latent space into several code vectors with an aim to capture one attribute in each of these vectors. Methods such as \cite{Mathieu2016DisentanglingFO,Szab2018ChallengesID, Jha2018DisentanglingFO} fall under this category but are limited to partitioning the latent space into two code vectors, such that a specified factor is captured in one, while all unspecified factors are captured in the other. More recently, \cite{Mixing_2018} utilized an autoencoder based mixing/unmixing approach to learn to disentangle in an unsupervised setting, while partitioning the latent space into several code vectors. While this approach allows to partition the latent space to several code vectors to capture multiple attributes, information of an attribute is not explicitly limited to only one code vector of the representation. 

While there is vast literature on disentanglement, it is still an open question to gauge the merit of approaches that impose variable independence over partitioned representations that dedicate multiple dimensions to capture one factor. In this work, we focus on improving the quality of the latter, by defining a prior structure on several code vectors learned by the network.
We undertake a first-principles based analysis, which leads us to a simple and elegant constraint on latent-spaces. The model is motivated by several independent results that provide interesting non-Euclidean structures to fundamental modes of image formation. These include models of lighting, deformation, pose, etc. which lead to latent-spaces as varied as the Grassmannian, the special orthogonal group, diffeomorphism groups etc. However, it turns out that if we are not too specific about {\em minimal} descriptions of such latent-factors, many of those geometric structures can actually be embedded into spheres of varying dimensions. The dimensions of the spheres are in turn related to the underlying dimension of the latent-factor under consideration. This realization leads us to develop a simplified model for the latent space where we assume a) spherical constraint per latent factor, b) equal dimensions per latent-block. These two assumptions lead to a rather simple analytical constraint in terms of an orthonormal constraint on latent-space blocks.  
More broadly, orthogonality constraint has been used as a regularization technique in many deep learning approaches. Various approaches like \cite{huang2018orthogonal}, have imposed orthogonality constraints on weights for efficient optimization and for stabilizing the distribution of activation maps in CNNs. The orthogonality constraint has also been utilized extensively in recurrent neural networks \cite{Vorontsov_RNNortho2017} to avoid gradient vanishing conditions. We show that a combination of both a soft constraint as well as imposing explicit orthogonality in the latent space while trianing improves the models' convergence, while also improving the quality of disentanglement. 

The contributions of this work are summarized as follows. a) We propose to parameterize the latent space representation as a product of orthogonal spheres, motivated by physical models of illumination, deformation, and motion. b) We simply the model to result in a simple ortho-normalization term in the loss function, and an update strategy that can be incorporated into most disentangling frameworks. c) We demonstrate both quantitatively and qualitatively that the proposed framework achieves improvement over state-of-the-art approaches on several benchmark datasets and metrics. 

\section{Related Work}\label{sec: pw}
Disentangling factors of variation to extract meaningful representations has been an engaging area of interest in the research community and has been explored in various contexts of style-transfer for both text \cite{John2018DisentangledRL} and image domains \cite{Gatys2016ImageST}, analogy-making \cite{Reed2015DeepVA} and domain adaptation \cite{Vu2017DomainAM, press2018emerging}. Most widely VAEs and GANs have been used to learn these representations in either unsupervised setting \cite{Chen2016InfoGANIR, Matthey2017betaVAELB, Kim2018DisentanglingBF, Li2018DisentangledSA, Chen2018IsolatingSO, press2018emerging, Shu2018DeformingAU} or semi/weakly supervised setting \cite{Mathieu2016DisentanglingFO, Jha2018DisentanglingFO, Szab2018ChallengesID, Bouchacourt2018MultiLevelVA, Ruiz2018LearningDR,Li2019}.

 Approaches such as those proposed in Mathieu \etal \cite{Mathieu2016DisentanglingFO} and Szab{\'o} \etal \cite{Szab2018ChallengesID} successfully separated a specified attribute using semi-supervised adversarial training. Ruiz \etal \cite{Ruiz2018LearningDR} also employ adversarial training in weak supervision. On the other hand, approaches like Jha \etal \cite{Jha2018DisentanglingFO} eliminates adversarial training and resort to cycle consistency in VAE framework to avoid degeneracy. The approach of \cite{Li2019} also uses cycle consistency, in a semi-supervised way, to disentangle pose and appearance for hand-pose tracking and estimation application. 
 
While these approaches perform well to some extent, disentangling in unsupervised setting is still a challenge. Several approaches like Press \etal \cite{press2018emerging} follow a domain adaptation approach using an encoder-decoder architecture to disentangle. Whereas, Li \etal \cite{Li2018DisentangledSA} utilized unsupervised sequence modeling to disentangle temporally dynamic features from the static ones in audio and video sequences. Another related approach, \cite{Shu2018DeformingAU} disentangles shape from appearance using canonical and template-based coordinate systems in an unsupervised manner.

Whether semi-supervised or unsupervised, most of these approaches are often limited to one or two factors of variation. In order to improve image synthesis and attribute transfer, a more elaborate representation is required that can disentangle multiple factors simultaneously.
More recently, there has been interest in disentangling multiple factors of variation simultaneously. Approaches like \cite{Choi2018StarGANUG} condition the input on a favorable mask of factors to be replicated in the output image. On the other hand, Bouchacourt \etal \cite{Bouchacourt2018MultiLevelVA}, utilized group level supervision, where within a group the observations share a common but unknown value for one of the factors of variation. This in effect relaxes the observations from independent and identically distributed (i.i.d) condition imposed in VAE based models, and also improves generalization to unseen groups during inference.

Unsupervised approaches based on $\beta$-VAE \cite{Matthey2017betaVAELB}, introduce a weighted KL-divergence term to ensure a restrained information bottleneck for the learned disentangled latent representations. $\beta$-TCVAE \cite{Chen2018IsolatingSO}, a variant of $\beta$-VAE, decomposes the marginal KL-divergence term into dimension-wise KL-divergence and a total correlation term that exhibits the extent of disentanglement. 
Another variant of $\beta$-VAE called Factor-VAE \cite{Kim2018DisentanglingBF}  supports the idea of the latent code distribution to be factorial by introducing a discriminator that differentiates a latent code picked from marginal code distribution from one that is a product of them.
Chen \etal \cite{Chen2016InfoGANIR} tries to achieve mutual information maximization between latent codes and corresponding observations via an additional information regularization in an unsupervised fashion. 
Although it works in an unsupervised fashion, lack of encoder-decoder style architecture limits the re-usability of the latent codes of an image. We aim to propose a general approach that works well on any disentangling model, by making use of factorized latent spaces to enhance the quality of disentanglement. 

More recently, over the last several years, geometry has been explored in deep learning  approaches. Approaches like \cite{Shao2018TheRG, Shukla2019GeometryOD} explored the Riemannian geometry of the latent space of generative models. \cite{Bang2018MGGANSM} incorporates a geometry-aware adversarial approach to avoid mode-collapse, while \cite{Li2018MRGANMR} adds a geometry-aware regularization in accordance with the manifold hypothesis. \cite{arvanitidis2018latent, Khnel2018LatentSN} consider the curvature and non-linear manifold statistics in latent-space and induce a Riemannian metric to achieve better interpolation and distance functions. \cite{Brahma2016WhyDL} proposes various metrics to measure disentanglement and flattening across layers of neural architectures.


\section{Proposed Approach}\label{sec: proposed}
 We would like to present a motivation for the choice of orthogonality in latent-spaces by first principles analysis of typical factors of image-formation. An image can be seen as a complex non-linear interaction between various factors such as lighting, pose, shape, and shading. In many cases, there is a wealth of literature that studies the basic geometric properties of each mode of variation. However, it has been found that the specific constraints are too specific for each factor, and generally not compatible with each other, or with contemporary machine learning models. In this section, we show that a slight relaxation of some of these classical findings will lead to a very natural compact model for latent-variables in the form of a product of orthogonal-spheres. This new model is very simple to implement and enforce as a simple loss-function in deep-learning modules. We start with a few common factors in image-formation, and develop the model.
\begin{enumerate}
    \item {\bf Lighting variables: } In the illumination cone model, assuming Lambertian reflectance, and convex object-shapes, one can show that the image space is a convex-cone in image space \cite{Georghiades2001}. A relaxation of this model leads to identifying cones as linear-subspaces, which are seen as points on a Grassmannian manifold $\mathcal{G}_{n,k}$ ($n$ = image-size, $k$ = lighting dimensions, typically considered equal to number of linearly independent normals on the object shape) \cite{Lohit2017}. Under certain conditions of variance on the Grassmannian being low, a distribution of points on the Grassmann induces a distribution on a high-dimensional sphere (see \cite{ChakrabortyICCV2015}), whose dimension depends on $n$ and $k$ \cite{ChakrabortyICCV2015}.  
    \item {\bf Pose variables: } 3D pose is frequently represented as an element of the special orthogonal group $SO(3)$. For analytical purposes, it is convenient to think of rotations represented by quaternions \cite{quaternions}, which are elements of the 3-sphere $S^3$ embedded in $\mathbb{R}^4$, with the additional constraint of antipodal equivalence. This makes rotations to be identified as points on a real-projective space $\mathbb{R}\mathbb{P}^3$. Real-projective spaces are just a special case of the Grassmannian -- in this case, of 1D sub-spaces in $\mathbb{R}^4$. Using similar result as before \cite{ChakrabortyICCV2015}, from a distribution on quaternions, we can induce a distribution in a higher-dimensional hyperspherical manifold.
    \item {\bf Deformation variables: } Deformation of underlying shape, or non-elastic deformation of image-grid (e.g. due to shape change, expression, or photoshop effects), can be modeled by the framework of diffeomorphic maps from $\mathbb{R}^2$ to $\mathbb{R}^2$. Diffeomorphic maps are continuous, smooth, and invertible maps that warp an image grid. Under additional conditions such as corner-point preservation, a square-root form of the diffeomorphic map can be viewed as a point on a infinite-dimensional Hilbert sphere \cite{Jermyn2012}. The infinite-dimensionality is more a mathematical convenience, but in practice, the dimension of the hypersphere is defined by the image-resolution. 
\end{enumerate}

\paragraph{Generality: } Now, the relevant question is whether hyperspherical relaxation is a good model for factors beyond the above? Consider for instance the factors listed in table \ref{table:map_celeba}. Most of the factors listed there are mid-level or semantic factors, rather than analytically definable physical factors. However, to a good degree of approximation such semantic factors can be explained as a result of combinations of low-level physical factors. Wavy hair for instance can be approximately described by deformation variables, makeup can be approximated by lighting variables, etc. While our model specifically is motivated by well-studied tractable factors like illumination, pose, and deformation, our conjecture is that the model is quite flexible and applicable beyond these factors. 
 
 \paragraph{Loss function: } Products of spheres is a generalization of tori geometry. The specification is incomplete without knowing the individual dimensions of the hyperspheres involved. For analytical and computational tractability, we choose to set the dimensions of the spheres to be equal to each other. This, in effect imposes a simple orthonormality constraint on the latent space blocks.

\begin{figure}[h]
\centering
\includegraphics[width = 0.8\textwidth,keepaspectratio]{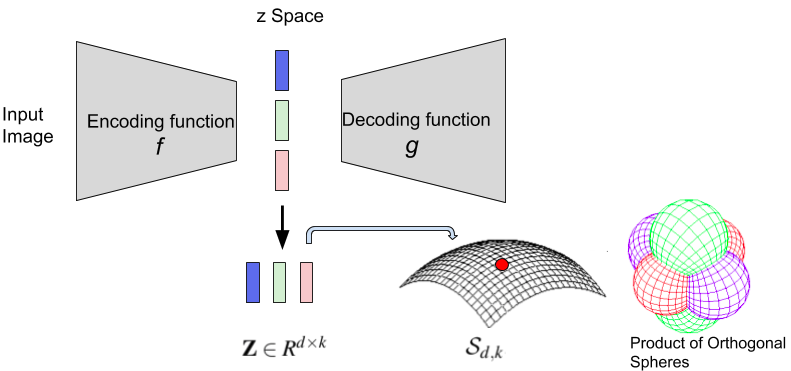}
   \caption{Illustration of proposed Product of Orthogonal Spheres as a latent space model}
\label{fig:model}
\end{figure}

Given a dataset $\calX = \{\bx_1, \bx_2, \ldots, \bx_m \} $, with $f$ and $g$ representing the encoding and decoding functions respectively such that $\bz = f(\bx)$ and $\bx = g(\bz)$. These functions can either represent a generative model or an autoencoder that is employed for disentangling. The latent space representation is given by $\bZ \in \bbR^{d\times k}=\{\bz^1, \bz^2, \ldots, \bz^k\}$, here $k$ is the number of attributes and $\bz^i \in \bbR^{d}$ represents code vector for each of the attributes. An illustration of the proposed parametrization and update strategy is shown in the Figure \ref{fig:model}. Under the PrOSe model, we would like to enforce orthonormality of $\bZ$ on an average across the entire training-data. That is, for the entire network, we augment the disentanglement loss function with a term penalizing deviations from orthonormality of the latent-space blocks, as:
\begin{align}
    \min_{\theta} \quad \calL_{disentangle} + \calL_{orth},
\end{align}
where, $\calL_{orth}$ is the ortho-normality constraint defined on the stacked matrix of the latent code vectors as columns in $\bZ \in \bbR^{d \times k}=[\bz^1 \bz^2  \ldots,  \bz^k]$, where $k$ is the number of latent code partitions and $d$ is the dimension of the latent code and is given by 
\begin{align}
    \calL_{orth} = \left\lVert \bZ^\top \bZ - \bI \right\rVert^2 _F,
\end{align}
where, $\left\lVert . \right\rVert_F$ is the matrix Frobenius norm. The loss-term includes a summation over training-samples in a given mini-batch, which is not explicitly noted above. 
\paragraph{Optimization Strategy.} We find that optimizing the cost function described above turns out to have slow convergence.
In order to speed up convergence, we add a more direct and explicit parameterization to the latent space. Now, the update strategy consists of two parts. Firstly, backpropagation works towards ensuring that the soft constraint is satisfied. And secondly, the hard constraint imposed by constraining the representation as a point on a Stiefel manifold i.e. $\calS_{d,k} =\{\bU \in \bbR^{d \times k}:\bU^\top\bU  = \bI\}$, helps in pushing the convergence along. We note that the hard constraint is only applied during training, and not at test-time. 
\\ \\
\textit{Optimizing the Stiefel Manifold Constraint:} Stiefel  manifold $\calS_{d,k}$ is defined as space of $d \times k$ matrices, that have orthonormal columns and is equipped with Frobenius inner product. Therefore, maintaining the orthonormality during the optimization is required. To this end, we utilize the Cayley transformation update that has been widely used in Stiefel manifold optimization techniques \cite{nishimori2005note,shukla2015distance}. The differentiability of this transformation allows the gradient information to backpropagate through this transformation. The update step is computed as follows.  Given an initial orthonormal matrix $\bZ$ and its Jacobian $\bJ$,  the Cayley update with a step size $\tau$ is given by:
\begin{align}
    \bZ_{new} = \big(\bI +\frac{\tau}{2}\bA\big)^{-1}(\bI - \frac{\tau}{2}\bA) \bZ,
\end{align}
where, $\bA = \bJ \bZ^\top - \bZ \bJ^\top$ is a skew symmetric matrix. The update step maps a skew-symmetric matrix to an orthogonal matrix using Cayley transform with learning rate $\tau$.

\section{Experiments} \label{sec: experiment}
This section highlights the quantitative and qualitative improvements achieved by the \textbf{PrOSe} model compared to state-of-the-art approaches. To evaluate the efficacy of our approach, we train our model on a diverse range of synthetic as well as real-world datasets.

\noindent \textbf{2D Sprites} \cite{Reed2015DeepVA} is a synthetic dataset with 143,040 images of animated game characters (sprites) with 480 distinct characters. The training set consists of 320 characters while validation and test sets contain 80 identities each. Annotations for distinct factors of variation such as gender, hair type, body type, armour type, arm type, and greaves type are provided. 

\noindent\textbf{MNIST} \cite{LeCun2001GradientbasedLA} comprises of 28$\times$28 grayscale images of hand written digits with 10 different classes. The training set consists of 60K images along with a test set of 10K images. A few tangible factors of variation that could be easily perceived from the data are stroke-width, slant angle, identity etc. 

\noindent \textbf{CelebA} \cite{liu2015faceattributes} is a celebrity face dataset with 202,599 each annotated with 40 partitionary attributes like eyeglasses, wearing hat, bangs, wavy hair, mustache, smile, oval face etc. The number of identities are 10,177. We trained our model using a train-test split ratio of 4:1. 
\\
\noindent\textbf{Implementation Details.} For 2D Sprites and CelebA dataset, the latent space is partitioned in 8 partitions with dimension 64 each. For MNIST, due to fewer variations in the data, the latent space is partitioned into 3 each with dimension 8. For MNIST, the choice is due to the fact that it consists of 3 prime attributes: class identity, stroke width and slant angle. The other two datasets have considerably more factors of variation and therefore, we aspire to capture the ones that are the most obviously perceptible. Additionally, the weights of individual loss terms used in our implementation is similar to \cite{Mixing_2018}. We use Adam optimizer with $\beta_{1}$ = 0.5 and $\beta_{2}$ = 0.999, and initial learning rate adapted to each dataset to ensure convergence. The step size ($\tau$) for Cayley transformation update equation is chosen as $0.1$ for all datasets.

\subsection{Results}
 We compare the efficacy of \textbf{PrOSe} framework with three approaches namely: MIX \cite{Mixing_2018}, $\beta$-VAE \cite{Matthey2017betaVAELB}, Factor-VAE \cite{Kim2018DisentanglingBF}. We integrate \textbf{PrOSe} framework with these and refer them as \textbf{PrOSe} + MIX, \textbf{PrOSe} + $\beta$-VAE and \textbf{PrOSe} + Factor-VAE.

\begin{wraptable}{r}{0.7\textwidth}
{\small 
\centering
	{\fontsize{7}{9}\selectfont
\begin{tabular}{|l|cc|cc|cc|}
\hline
Attribute & MIX & +\textbf{PrOSe} & $\beta$-VAE & +\textbf{PrOSe} & Factor-VAE & + \textbf{PrOSe}\\
\hline\hline
Eyebrows & 79.4 & 79.5 & 78.8 & 79.4 & 79.4 & 79.8\\
Attractive & 72.6 & 80.4 & 73.6  & 74.6 & 74.4 & 76.2\\
Bangs & 91.7 & 90.2 & 92.6 & 92.8 & 93.2 & 94.2\\
Black Hair & 71.9 & 75.6 & 72.4 & 78.2 & 76.5 & 78.1\\
Blonde Hair & 87.2 & 92.0 & 88.4 & 90.2 & 88.2 & 88.4\\
Makeup & 76.5 & 78.0 & 76.2 & 77.8 & 77.1 & 77.5\\
Male & 86.2 & 83.1 & 84.6 & 86.4 & 83.8 & 83.2\\
Mouth & 72.0 & 80.6 & 73.8 & 74.1 & 74.4 & 75.0\\
No Beard & 86.3 & 89.6 & 86.0 & 86.3 & 85.2 & 85.8\\
Wavy Hair & 65.7 & 71.9 & 66.2 & 67.0 & 66.8 & 67.0\\
Hat & 95.2 & 94.8 & 93.5 & 95.4 & 93.0 & 93.8\\
Lipstick & 79.8 & 80.5 & 82.4 & 83.2 & 83.3 & 87.5\\
\hline
\hline
Average & 80.3 & \textbf{83.0} & 80.7 & {\bf 82.1} & 81.3  & {\bf 82.2}\\
\hline
\end{tabular}
}
\caption{mAP values for different attributes for CelebA face dataset with various approaches with and without \textbf{PrOSe}  parameterization.}
\label{table:map_celeba}
}
\end{wraptable}

\noindent \textbf{Classification Performance.} We report Mean Average Precision (mAP) values to quantify the classification, with higher mAP indicating better performance.  Table \ref{table:map_celeba} and \ref{table:map_sprites} show improved results with PrOSe for 2D Sprites  and CelebA dataset.
 \begin{wraptable}{r}{0.7\textwidth}
{\small 
\centering
	{\fontsize{8}{9}\selectfont
\begin{tabular}{|l|c|c|c|c|c|c|c|}
\hline
Method & Body & Skin & Vest & Hair & Arm & Leg & Avg.\\
\hline\hline
MIX  & 66.5 & 77.2 & 90.0 & 56.2 & 63.1 & 89.4 & 73.7\\
+ \textbf{PrOSe} & 70.1 & 75.5 & 88.5 & 63.2 & 72.1 & 94.4 & {\bf 77.3}\\\hline\hline
$\beta$-VAE  & 66.8 & 78.4 & 89.7 & 57.2 & 63.3 & 90.1 & 74.0\\
+ \textbf{PrOSe}& 70.2 & 77.3 & 88.1 & 65.3 & 73.0 & 93.7 & \textbf{77.7}\\\hline \hline
Factor-VAE  & 66.7 & 78.6 & 90.3 & 57.3 & 63.1 & 90.1 & 74.4\\
+ \textbf{PrOSe} & 69.8 & 77.1 & 89.0 & 63.8 & 72.7 & 93.7 & \textbf{77.7}\\
\hline
\end{tabular}
}
\caption{mAP values for different attributes for 2D Sprites with various approaches with and without \textbf{PrOSe}.}
\label{table:map_sprites}
}
\end{wraptable}
 
\begin{figure}[h!]
\centering
\subfigure[MNIST: Identity (Pair 1), Slant Angle (Pair 2) and Stroke Width (Pair 3)]{\includegraphics[width=2cm]{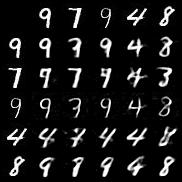} 
\includegraphics[width=2cm]{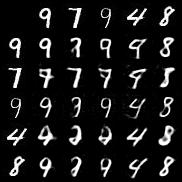}
\hspace{0.17cm}
\includegraphics[width=2cm]{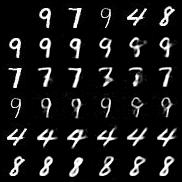}
\includegraphics[width=2cm]{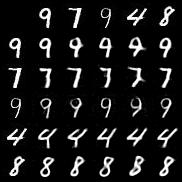}
\hspace{0.17cm}
\includegraphics[width=2cm]{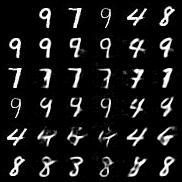}
\includegraphics[width=2cm]{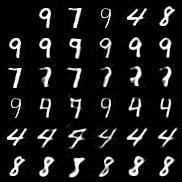}}
\subfigure[2D Sprites: Complexion (Pair 1), Hair colour (Pair 2) and armour colour (Pair 3)]{
\includegraphics[width=2cm]{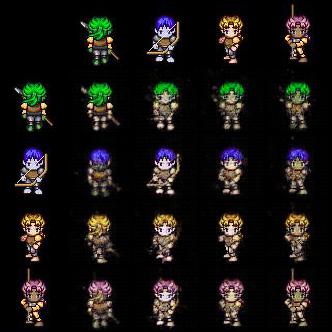}
\includegraphics[width=2cm]{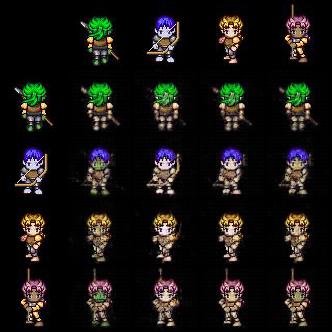}
\hspace{0.1cm}
\includegraphics[width=2cm]{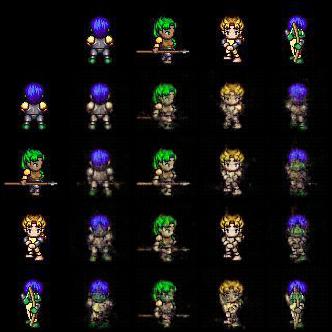}
\includegraphics[width=2cm]{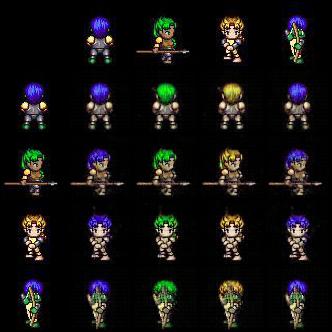}
\hspace{0.1cm}
\includegraphics[width=2cm]{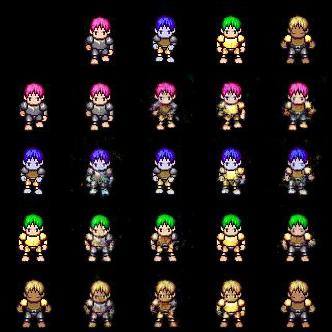}
\includegraphics[width=2cm]{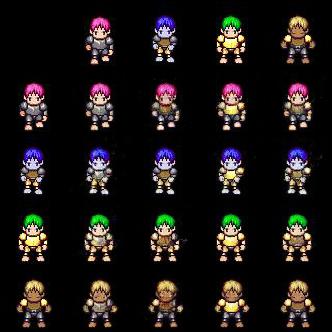}}
\subfigure[CelebA: Moustache (Pair 1), Complexion (Pair 2) and Hair (Pair 3)]{\includegraphics[width=2cm]{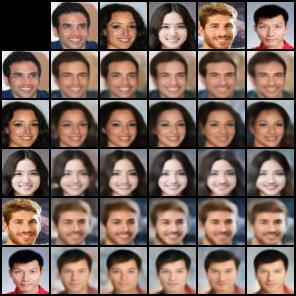}
\includegraphics[width=2cm]{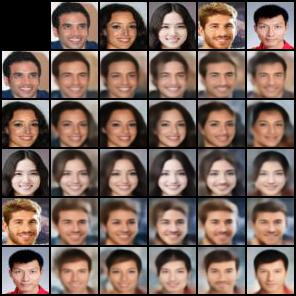}
\hspace{0.1cm}
\includegraphics[width=2cm]{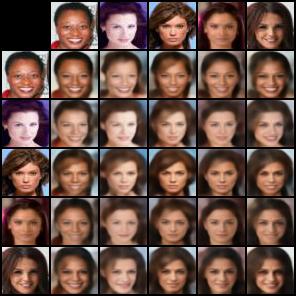}
\includegraphics[width=2cm]{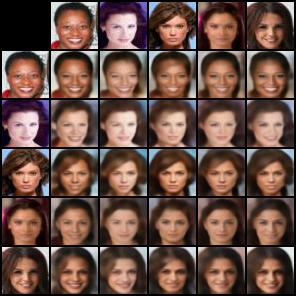}
\hspace{0.1cm}
\includegraphics[width=2cm]{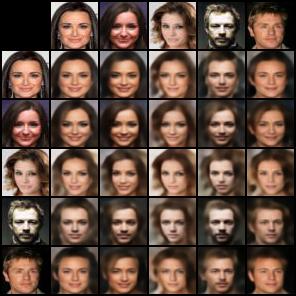}
\includegraphics[width=2cm]{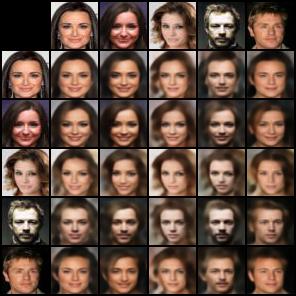}}
    \caption{A visualization grid of image synthesis using attribute transfer. In each grid of the subfigures, the top row and leftmost column images come from the test set. The other images are generated  using  code vector corresponding to one of the attribute from the image in this top row, while all remaining attributes are taken from the leftmost column image. Results with MIX(left in each pair of the subfigure) and \textbf{PrOSe} (right in each pair) are shown. } 
\label{fig:comb_grid}
\end{figure}

\noindent\textbf{Attribute Transfer and Image Synthesis.}
Figure \ref{fig:comb_grid}  shows  attribute transfer to visualize the quality of disentanglement. The first row and the first column in each grid are randomly chosen samples from the test-set. All other images are formed by picking one attribute from the corresponding column lead and remaining attribute code vectors are kept intact as those of latent space embeddings of the corresponding row lead. This allows to quantify, how well an attribute is captured in a single partition. 
For example, in 2D Sprites, more than one factor changes simultaneously with MIX \cite{Mixing_2018}. i.e. along with complexion/hair colour, a variation in pose is seen, similarly, leg colour also varies along with armour colour. Similar trend is observed in other datasets, where either multiple factors change at or the specified attribute is not transferred. 

\begin{figure}[h!]
\begin{center}
\includegraphics[width=10cm]{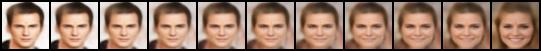}
\includegraphics[width=10cm]{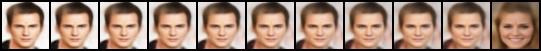}
\end{center}
\vspace{-0.6cm}
   \caption{Interpolation across disentangled gender attribute for CelebA datatset with MIX (top) and with \textbf{PrOSe}  + MIX (bottom). \textbf{PrOSe} achieves a well separated attribute spaces, evidenced by smoother and more meaningful interpolation without altering face shape, expressions etc.}
\label{fig:interpolation_celebA}
\end{figure}

\begin{figure}[h!]
\begin{center}
\includegraphics[width=6cm]{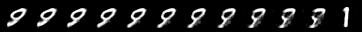}
\includegraphics[width=6cm]{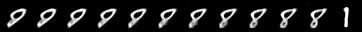}
\includegraphics[width=6cm]{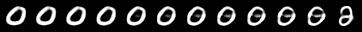}
\includegraphics[width=6cm]{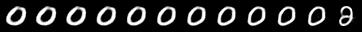}
\includegraphics[width=6cm]{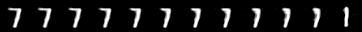}
\includegraphics[width=6cm]{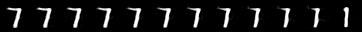}
\\
\includegraphics[width=6.3cm]{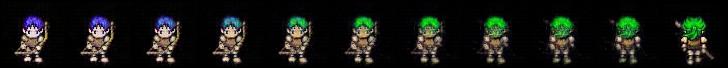}
\includegraphics[width=6.3cm]{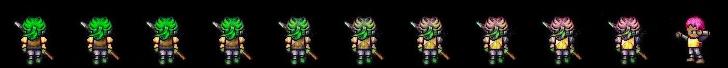}
\includegraphics[width=6.3cm]{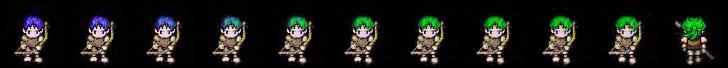}
\includegraphics[width=6.3cm]{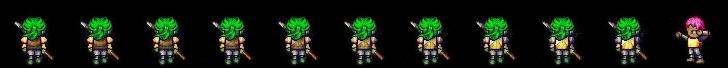}
\end{center}
\vspace{-0.6cm}
   \caption{Interpolation across disentangled individual partitions signifying different attributes: For MNIST Angle (Row 1), Stroke Width (Row 2), Identity (Row 3), For Sprites ; Hair (Row 4) and Armour (Row 5) with MIX (Left) and \textbf{PrOSe} +MIX (Right).}
\label{fig:interpolation_mnist_sprites}
\end{figure}
\begin{figure}[h]
\centering
\includegraphics[width=4cm,trim={1.5cm 2cm 1cm 0.5cm}]{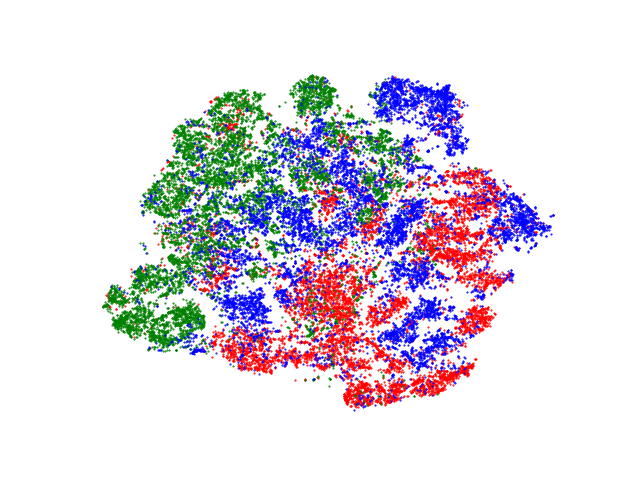}
\includegraphics[width=4cm,trim={1.5cm 2cm 1cm 0.3cm}]{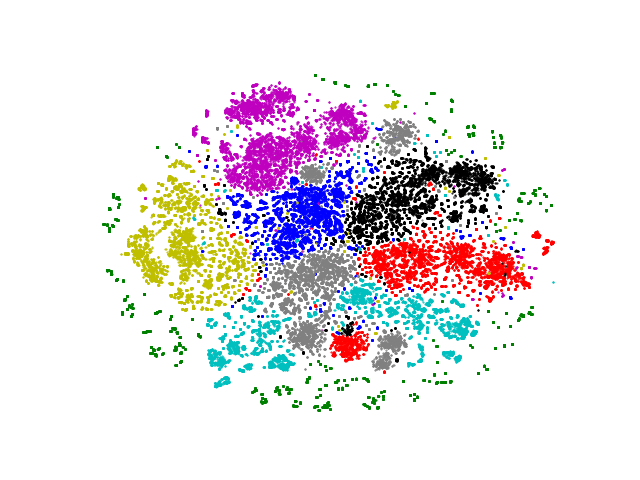}
\includegraphics[width=4cm,trim={1.5cm 2cm 1cm 0.3cm}]{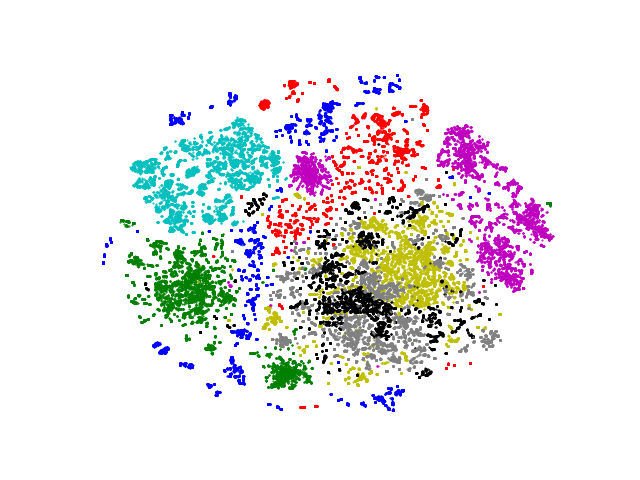}
\includegraphics[width=4cm,trim={1.5cm 2cm 1cm 0.3cm}]{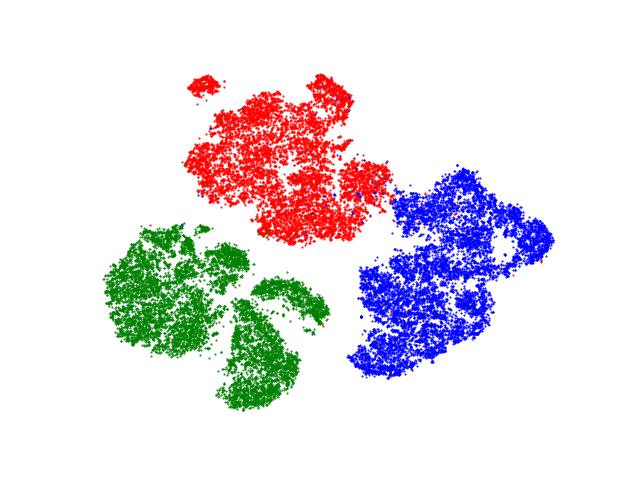}
\includegraphics[width=4cm,trim={1.5cm 2cm 1cm 0.3cm}]{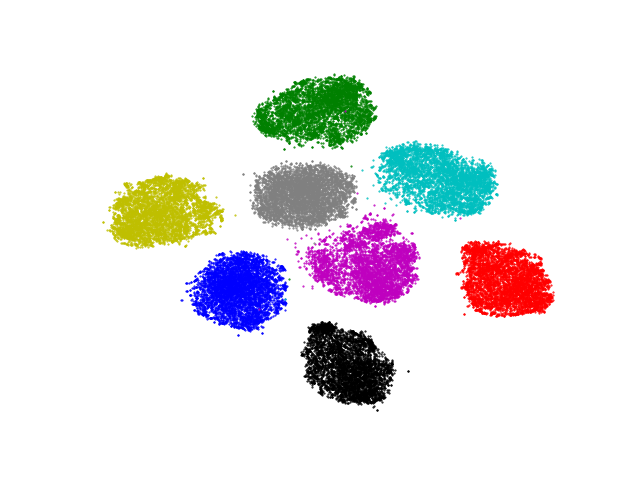}
\includegraphics[width=4cm,trim={1.5cm 2cm 1cm 0.3cm}]{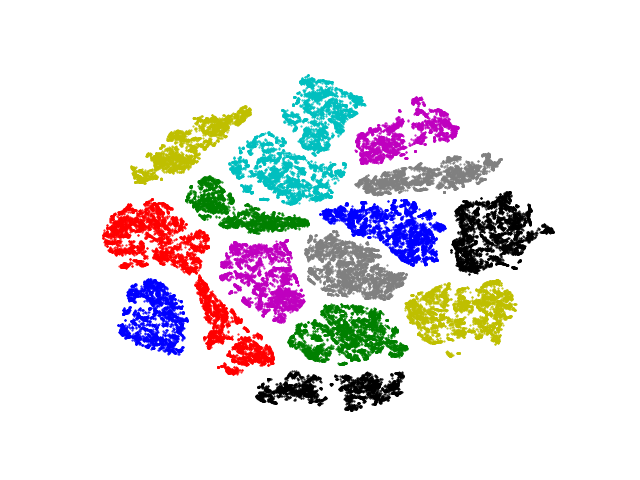}
   \caption{t-SNE Plots for MNIST (Column 1), 2D Sprites (Column 2) and CelebA face (Column 3) datasets with MIX (top) and \textbf{PrOSe} +MIX (bottom). The different colors denote different attribute spaces. Clearer separation of attributes is seen in the case of {\bf PrOSe}.}
\label{fig:tsne}
\end{figure}
\begin{figure}[h!]
\begin{center}
\vspace{-0.1cm}
\subfigure[MNIST]{\includegraphics[width=2cm]{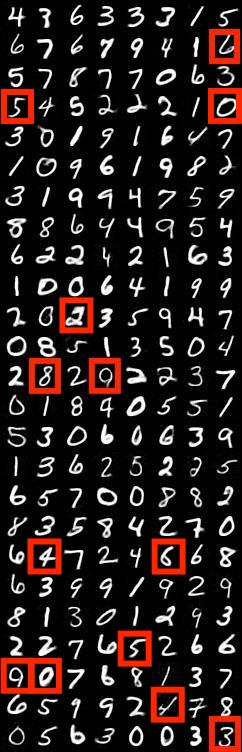}
\includegraphics[width=2cm]{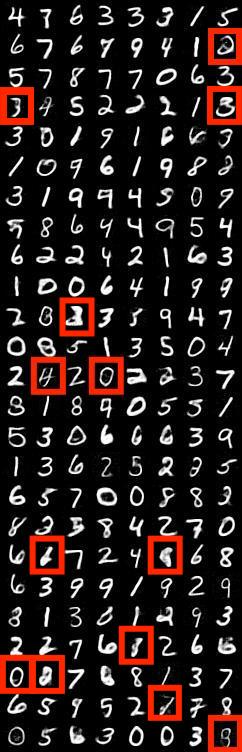}
\hspace{2cm}
\includegraphics[width=2cm]{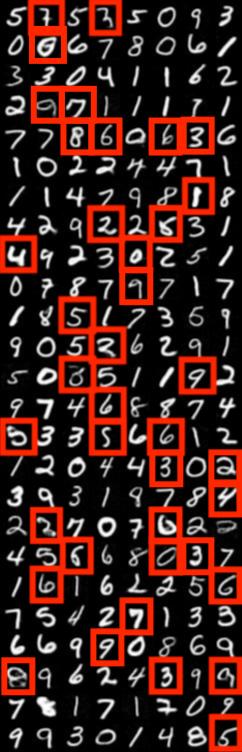}
\includegraphics[width=2cm]{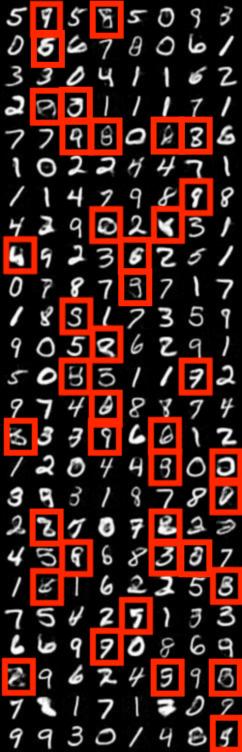}}
\subfigure[2D Sprites]{\includegraphics[width=3cm]{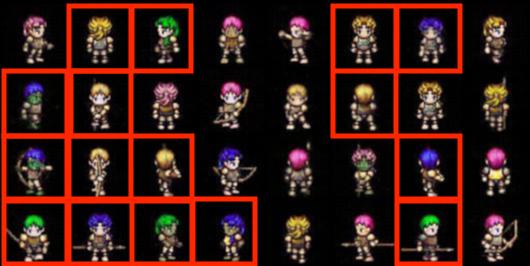}
\includegraphics[width=3cm]{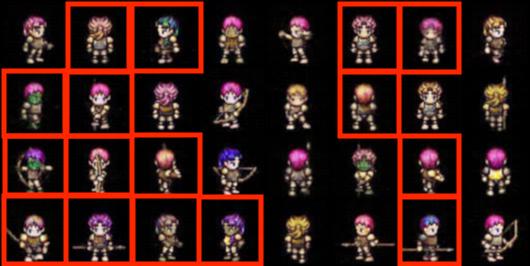}
\hspace{0.5cm}
\includegraphics[width=3cm]{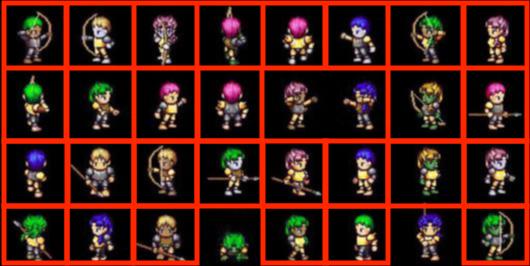}
\includegraphics[width=3cm]{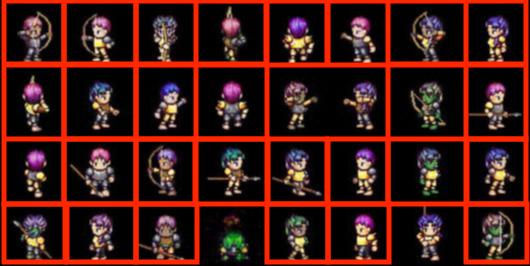}}

\end{center}
   \caption{Results of predicting identity (for MNIST) and hair colour (for 2D Sprites) using remaining attributes with MIX \cite{Mixing_2018} and \textbf{PrOSe} + MIX. In both cases, (Left) shows true class and predicted class by MIX, (Right) shows true class and predicted class by \textbf{PrOSe} + MIX. Marking on the true class is shown for visual correspondence. The marked red boxes are the misidentified images.  \textbf{PrOSe} + MIX results in more mis-classifications, indicating that disentanglement is more effectively occurring.}
\label{fig:prediction_model}
\vspace{-0.5cm}
\end{figure}

\noindent \textbf{Interpolation.} Interpolation between two images, along one of the attributes while fixing others, demonstrates the purity of the code vectors. Figure \ref{fig:interpolation_celebA} and \ref{fig:interpolation_mnist_sprites} show that \textbf{PrOSe} achieve a smoother transition in the attribute of interest while others are unaltered. For example, in CelebA, interpolating in gender attribute space, also affects smile and even face shape, in the case of MIX \cite{Mixing_2018}. The smooth variations within an attribute space can also be established with the low dimensional projections of different attributes in Figure \ref{fig:tsne}. It can be observed that \textbf{PrOSe} achieves a well separated attribute space, evidenced by ensuring smoother and more meaningful interpolation without altering face shape, expression etc.

\noindent\textbf{Prediction Model.}
To validate the quality of disentanglement with PrOSe, we trained a model that uses a random set of $k - 1$ attributes ($k$ being the total number of partitions in the latent space) to predict the remaining attribute. 
In an ideal scenario, the prediction model must fail, since the left-out attributes information is not captured by the other partitions. Thus, a higher number of mis-classification signifies lesser information leakage in the learned partitioned representations. Figure \ref{fig:prediction_model} shows increased error with PrOSe framework. Each red marking in the figure indicates a mis-prediction.

\noindent\textbf{Generalizability.} PrOSe is also evaluated with 
$\beta$-VAE \cite{Matthey2017betaVAELB} and Factor-VAE \cite{Kim2018DisentanglingBF} that encode different factors of variation along the different dimensions of latent space. To evaluate the \textbf{PrOSe} framework, the 12 dimensional latent space is partitioned into 3 equal sized partitions of size 4 each. The mAP results are given in Table \ref{table:map_celeba} and \ref{table:map_sprites}. Due to space constraints, the interpolation results are presented in the supplementary material. 
We find that PrOSe reduces information leakage among different partitions while maintaining good image quality. 

\section{Conclusion}
In this work,  we used a product of orthogonal spheres parameterization for learning improved disentangled representations. The model amounts to imposing an orthonormality constraint on the partitioned latent representation of a data sample. The proposed framework, requires adding a simple loss term to the disentangling loss, making it easy to incorporate into state-of-the-art approaches.  We show that the proposed parameterization successfully improves the quality of disentanglement in various datasets. Directions for future work include investigating broadening the model to different-sized latent blocks, and the impact of the model on other tasks such as recognition, multi-task learning, and generalization to other unseen factors.

\section*{Acknowledgements}
This work was supported partially by ARO grant number W911NF-17-1-0293, at Arizona State University, USA  and Infosys Center for Artificial Intelligence at IIIT- Delhi, India.
\bibliography{Main}
\end{document}